\documentclass{article}

\usepackage{arxiv}

\usepackage[utf8]{inputenc} 
\usepackage[T1]{fontenc}    
\usepackage{hyperref}       
\usepackage{url}            
\usepackage{booktabs}       
\usepackage{amsfonts}       
\usepackage{nicefrac}       
\usepackage{microtype}      
\usepackage{lipsum}		
\usepackage{graphicx}
\usepackage{doi}
\usepackage{placeins} 
\usepackage{amsmath}
\usepackage[dvipsnames]{xcolor}
\usepackage[ruled, linesnumbered]{algorithm2e}

\title{A Search for Nonlinear Balanced Boolean Functions by
Leveraging Phenotypic Properties}

\author{ \href{https://orcid.org/0000-0003-0486-233X}{\includegraphics[scale=0.06]{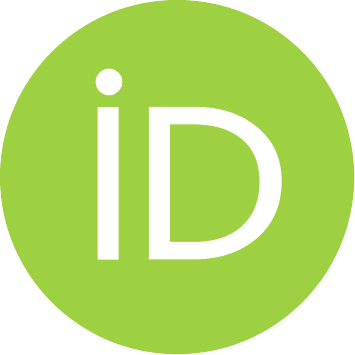}\hspace{1mm}Bruno Gašperov} \\
        University of Zagreb, Faculty of \\
        Electrical Engineering and Computing \\
        Zagreb, Croatia \\
	\texttt{bruno.gasperov@fer.hr} \\
	\And
	\href{https://orcid.org/0000-0001-8732-4769}{\includegraphics[scale=0.06]{orcid.pdf}\hspace{1mm}Marko Đurasević} \\
        University of Zagreb, Faculty of \\
        Electrical Engineering and Computing \\
        Zagreb, Croatia \\
	\texttt{marko.durasevic@fer.hr} \\
 	\And
	\href{https://orcid.org/0000-0002-9201-2994}   {\includegraphics[scale=0.06]       {orcid.pdf}\hspace{1mm}Domagoj Jakobović} \\
        University of Zagreb, Faculty of \\
        Electrical Engineering and Computing \\
        Zagreb, Croatia \\
	\texttt{domagoj.jakobovic@fer.hr} \\
}

\date{}

\renewcommand{\shorttitle}{A Search for Nonlinear Balanced Boolean Functions by
Leveraging Phenotypic Properties}

\hypersetup{
pdftitle={A Search for Nonlinear Balanced Boolean Functions by
Leveraging Phenotypic Properties},
pdfsubject={q-bio.NC, q-bio.QM},
pdfauthor={David S.~Hippocampus, Elias D.~Striatum},
pdfkeywords={First keyword, Second keyword, More},
}

\begin{document}
\maketitle

\begin{abstract}
In this paper, we consider the problem of finding perfectly balanced Boolean functions with high non-linearity values. Such functions have extensive applications in domains such as cryptography and error-correcting coding theory. We provide an approach for finding such functions by a local search method that exploits the structure of the underlying problem. Previous attempts in this vein typically focused on using the properties of the fitness landscape to guide the search. We opt for a different path in which we leverage the phenotype landscape (the mapping from genotypes to phenotypes) instead. In the context of the underlying problem, the phenotypes are represented by Walsh-Hadamard spectra of the candidate solutions (Boolean functions). We propose a novel selection criterion, under which the phenotypes are compared directly, and test whether its use increases the convergence speed (measured by the number of required spectra calculations) when compared to a competitive fitness function used in the literature. The results reveal promising convergence speed improvements for Boolean functions of sizes $N=6$ to $N=9$.
\end{abstract}

\keywords{Boolean functions, local search, phenotype landscape, Walsh-Hadamard Transform, evolutionary computing, mutation}

\section{Introduction}
Boolean functions find their applications in a wide range of areas, such as cryptography and error-correcting codes \cite{carlet2010boolean}, telecommunications \cite{paterson2004dentist}, systems biology \cite{wang2012boolean}, and circuit design \cite{al2020distributed}. The properties of Boolean functions that are of critical importance, especially in cryptography, include high non-linearity and balancedness. Non-linearity indicates its distance from the closest affine function, while balancedness points to the equality of the number of zeros and ones in its truth table. In order to find Boolean functions with desired characteristics, metaheuristics such as genetic algorithms (GAs), local search, and algebraic constructions, are commonly used \cite{picek2016evolutionary, picek2016maximal, zhang2014generalized}. While balancedness is trivial to check, calculation of non-linearity involves calculating the Walsh-Hadamard spectrum, which is computationally costly even when the fast Walsh-Hadamard transform algorithm is used\footnote{The fast Walsh-Hadamard transform algorithm reduces the complexity from  $\mathcal{O}(N^2)$ (naive implementation) to $\mathcal{O}(N\log{}N)$.} \cite{fino1976unified}. This is especially problematic for Boolean functions with larger numbers of inputs $N$. On top of that, the size of the search space itself equals $2^{2^{N}}$, i.e., it grows super-exponentially with $N$. Hence, reducing the number of Walsh-Hadamard spectrum evaluations is pivotal to increasing the convergence speed of practically any approach for finding Boolean functions with high non-linearity based on metaheuristics.

To this end, we propose a novel selection criterion for use in metaheuristics which is then employed to directly compare the phenotypes of candidate solutions (Boolean function) and ascertain which solution is more likely to have neighbors with higher non-linearity values. The underlying idea is to exploit the structure of the phenotype landscape - the mapping between the genotypes (Boolean functions encoded as bitstrings) and phenotypes (Walsh-Hadamard spectra). More specifically, the design of the criterion is driven by the close links between phenotypic and fitness landscapes for the underlying problem, given that fitness is typically defined as equal to the non-linearity, which is in turn related to the maximum absolute value of the Walsh-Hadamard spectrum (phenotype). Hence, our work contributes to exploring the relationship between the two landscapes. To demonstrate its viability, the selection criterion is incorporated into a simple first-improvement local search algorithm. It should be noted that our work builds upon the nascent strand of related research in which the goal is to somehow unveil the (in the context of this problem very intricate and convoluted) structure of the phenotype landscape \cite{djurasevic2023digging}. This area of research is still very scarce, as most works, in contrast, focus on exploiting or extensively analyzing the structure of the classical fitness landscape under simple fitness functions (e.g. non-linearity) \cite{jakobovic2021toward}. Another aspect in which our work differs is its exclusive focus on (perfectly) balanced Boolean functions, to which somewhat less attention has been paid in the literature. Interestingly, despite the fact that the consideration of only balanced functions significantly shrinks the search space, this in practice does not lead to noticeable improvements when using metaheuristic-based approaches, such as genetic algorithms. A presumed reason is the increased roughness of the ensuing fitness landscape \cite{manzoni2022influence}.

The paper is organized as follows. In Section \ref{chapter:2}, preliminaries on Boolean functions and the concept of a phenotype landscape are given. Related work is discussed in Section \ref{chapter:25}. Section \ref{chapter:4} presents the first-improvement local search strategy, the proposed phenotype-based selection criterion, and other key components of the method. The experimental results are given in Section \ref{chapter:5}, including first an introductory analysis of the relationship between different types of balancedness-preserving mutation operators and the non-linearity values of the resulting neighbors and then the main results obtained via two local search variants. Finally, Section \ref{chapter:6} wraps the paper up with a conclusion and a list of potential avenues for further research.

\section{Preliminaries}
\label{chapter:2}
\subsection{Boolean functions}
\label{chapter:2.1}
A Boolean function is a function in which the input and function values assume only two values, namely 0 and 1. 
We define an $n$-variable Boolean function as a mapping $f: \mathbb F_2^N \to \mathbb F_2$.
Each such function can be uniquely represented by a truth table, which represents pairs of inputs $x \in \mathbb F_2^N$ and function values $f(x)$ corresponding to those inputs. 
The vector of all output values $f(x)$ is called the value vector $\Omega_f$.
The size of this vector is $2^N$, whereas the size of the search space is equal to $2^{2^N}$. 
A common requirement for Boolean functions is that they should have the highest possible non-linearity.
The non-linearity ($nl_f$) of a Boolean function is defined as the minimum Hamming distance between a Boolean function and all affine functions, where the Hamming distance between two functions $f$ and $g$ denotes the number of different output values for the same input value, i.e. the number of inputs $x \in F_2^N$ such that $f(x)\neq g(x)$.
For a given Boolean function, we can calculate its non-linearity as: 
\begin{equation}
nl_{f} = 2^{N - 1} - \frac{1}{2}\max_{a \in \mathbb F_{2}^{N}} |W_{f}(a)|,
\end{equation}
where $W_f(a)$ represents the Walsh-Hadamard coefficient. These coefficients can be calculated using the Walsh-Hadamard transform defined as:
\begin{equation}
W_{f} (a) = \sum\limits_{x \in \mathbb F_{2}^{N}} (-1)^{f(x) \oplus a\cdot x},
\end{equation}
where $\oplus$ denotes the addition modulo two (bitwise XOR) and $a\cdot x$ denotes the logical AND of $a$ with each coordinate of $x$. This expression measures the correlation between the function $f$ and the linear function $a \cdot x$. As previously stated, a common goal is to obtain Boolean functions of the highest possible non-linearity value, which happens when the maximum absolute value of the corresponding Walsh-Hadamard is as small as possible. 
One interesting variant of Boolean functions is provided by balanced Boolean functions. These functions have a Hamming weight of $2^{N-1}$, meaning that the value vector consists of an equal number of zeros and ones. Balanced Boolean functions are of particular interest since they are more appropriate for being used in cryptosystems \cite{Carlet2007}, due to them not having a bias in their output values.

\subsection{Phenotype landscape}
Given the set of genotypes $\Theta$ and the set of phenotypes $\mathcal{P}$ (typically $\mathcal{P} \subseteq \mathbb{R}^n$), a phenotype function $p: \Theta \mapsto \mathcal{P}$ simply maps genotypes (parameters) to phenotypes:
\begin{equation}
p(\boldsymbol{\theta}_i) = \boldsymbol{p}_i,
\end{equation}
where $\boldsymbol{\theta}_i$ is any genotype and $\boldsymbol{p}_i$ its uniquely defined phenotype. In our case, $\boldsymbol{\theta}_i$ corresponds to a bitstring representation of a perfectly balanced Boolean function and $\boldsymbol{p}_i$ to its Walsh-Hadamard spectrum. The landscape associated with the mapping $p$ is referred to as the phenotype landscape. As a side note, we mention that in cases when phenotypes represent behaviors of controlled agents and genotypes stand for parameters of the associated controller, which is a common scenario in evolutionary reinforcement learning, this type of landscape is also referred to as a behavior (or feature) landscape.

\section{Related work}
\label{chapter:25}
There is a rich history of work in finding highly non-linear Boolean functions via metaheuristics \cite{djurasevic2023survey}, starting with Millan \emph{et. al.} \cite{millan1997effective, millan1998heuristic}, who approached the problem with a basic GA, a directed hill climbing method, and also a GA with hill climbing. Other attempts use alternative algorithms like particle swarm optimization \cite{mariot2015heuristic}, simulated annealing \cite{clark2000two}, or the clonal selection algorithm (CLONALG) \cite{PICEK2017320}. Some researchers, like Manzoni \emph{et al.} \cite{manzoni2022influence}, incorporated different local search methods into metaheuristics to improve and study the resulting convergence speed and diversity. Throughout the years, various solution representations have been proposed when dealing with the construction of Boolean functions.
The most natural and commonly used representation is the bitstring representation, in which the truth table is encoded as a string of bits of length $2^N$. However, recent years saw a rise in the popularity of symbolic-based representations, in which the Boolean function is defined as an expression that can be executed.
Various algorithms relying on such a representation have been considered in the literature, such as genetic programming \cite{10.1145/2464576.2464671} and its Cartesian variant \cite{10.1007/978-3-319-16501-1_16}. 
Although this representation was found to be the most successful one, it still lags behind the bitstring representation in popularity \cite{djurasevic2023survey}. Apart from the aforementioned representation, other representations were also proposed, such as the integer-based \cite{picek18a} and floating point representations \cite{PICEK2017320}, but neither received a lot of attention in the literature.
An alternative approach to representing Boolean functions is to use the Walsh-Hadamard spectrum to encode solutions \cite{Clark2003}. Although it is a promising approach, its performance is inferior in comparison to the traditional truth table-based representations.

Until now, non-linearity has been the most commonly considered criterion when evolving Boolean functions, especially in the context of single objective optimization \cite{FullerDM03, ManzoniMT21}. 
However, there have been several attempts to construct Boolean functions while considering multiple criteria simultaneously, such as optimizing non-linearity together with algebraic degree \cite{10.1007/978-3-319-26841-5_6} or autocorrelation \cite{10.1145/2739482.2764674}. 
Some studies even consider optimizing more than two criteria simultaneously \cite{cryptoeprint:2013:011, 10.1007/978-3-319-10762-2_80}, with good results being achieved.
Most studies dealing with the simultaneous optimization of multiple criteria use either a linear weighted combination of the criteria or a two-level approach in which the first criterion is optimized until a desired value is reached, after which the second one is optimized.
However, certain studies consider the application of multi-objective algorithms for this purpose as well \cite{10.1145/1276958.1277112}. 

Only a handful of studies have focused on analyzing the evolutionary process in order to gain deeper insight into the search for high-quality Boolean functions. 
Among works focusing on fitness landscape analysis, Jakobovic \emph{et al.} \cite{jakobovic2021toward} rely on local optima networks (LONs) in order to investigate the influence of different decisions (fitness functions, neighborhood operators, etc.) on the optimization of cryptographic properties. 
Somewhat similarly, Picek \emph{et al.} \cite{picek2015fighting} complement fitness analysis with a symmetry analysis, noting that this additional information might lead to more effective search methods.
Furthermore, Picek \emph{et al.} \cite{Picek2016} also conduct a fitness landscape analysis in order to analyze the difficulty in obtaining maximal possible non-linearity when constructing balanced Boolean functions.
The performed fitness landscape analysis did not reveal any differences in the landscapes between problems of different input sizes that could justify the increase in the problem difficulty when the number of input variables is increased.

\section{Method}
\label{chapter:4}

\subsection{Genotypes and phenotypes}

The genotypes are simply given as truth tables in the bitstring representation. The corresponding phenotypes are naturally provided by its Walsh-Hadamard (magnitude) spectrum, and this formulation is therefore used in our approach. Alternatively, phenotypes might be defined, for example, as a set of statistics of the spectrum. 

\subsection{Variation operators}

We consider several types of balancedness-preserving mutation operators: a) swaps (both single and multiple), b) cyclic shifts, c) inversions, and d) permutations. The analysis will be used to inform the choice of the operator used in the ensuing local search algorithm. We consider the effect of the three main types of mutations on multiple fitness functions and differently defined phenotypes. Let us denote a binary string by $s$, where $s=s_{1} s_{2} \ldots s_{L}$ where $s_i$ is its $i$-th element, and $L$ its length. In our case, $s=\Omega_f$ for some Boolean function $f$, and $L=2^{N}$, using the same notation as in Subsection \ref{chapter:2.1}. Given two binary strings $s$ and $t$, the Hamming
distance $\mathcal{H}(s, t)$ is simply defined as the number of elements (positions) in which they differ.

In a single \textbf{swap}, two indices $i$ and $j$, $i \neq j$, such that $s_{i} \neq s_{j}$, are randomly selected and the corresponding values are swapped to obtain a new binary string $s'$. In a multiple swap, the same procedure is repeated multiple ($k$) times, all while ensuring that subsequent swaps do not undo previous ones. \textbf{Cyclic shifts} simply move the elements of the string to the right by $l$ positions, while those that "fall off" the end are then re-added to the beginning of the string. Finally, in an \textbf{inversion}, two indices $i$ and $j$, $i < j$, are randomly chosen and the values in the substring $s_{i} \ldots s_{j}$ are inverted (replaced by the substring $s_{j} \ldots s_{i}$), and hence $s' = s_1 \ldots s_{i-1} s_{j} \ldots s_{i} s_{j+1} \ldots s_L$, where $s'$ again denotes the new (mutated) string. If $s' = s$ the process is repeated until $s' \neq s$. Finally, a \textbf{permutation} is given by a generalization of an inversion. Again, two indices $i$ and $j$, $i < j$, are randomly chosen and the values in the substring $s_i \ldots s_j$ are permuted such that $s' \neq s$ is ensured. Note that, in this case, the neighborhood of any $s$ equals the space of all perfectly balanced Boolean functions. The effective sizes of the neighborhoods associated with each mutation type and concrete examples for $L=6$ are given in Table \ref{tab:freq}. 

\begin{table}
  \caption{Types of balancedness-preserving mutations together with upper bounds on the effective neighborhood size. The affected positions in each example are colored.}
  \label{tab:freq}
  \begin{center}
  \begin{tabular}{ccl}
    \toprule
    Mutation&Example&Neigh. size\\
    \midrule
    swap (single) & $[\textcolor{blue}{1},\textcolor{violet}{0},0,1,0,1] \rightarrow [\textcolor{violet}{0},\textcolor{blue}{1},0,1,0,1]$ & $< L(L-1)/2$\\
    cyclic shift & $[\textcolor{blue}{1},\textcolor{violet}{0},\textcolor{orange}{0},\textcolor{teal}{1},\textcolor{magenta}{0},\textcolor{cyan}{1}] \rightarrow [\textcolor{cyan}{1},\textcolor{blue}{1},\textcolor{violet}{0},\textcolor{orange}{0},\textcolor{teal}{1},\textcolor{magenta}{0}]$  & $<L-1$\\
    inversion & $[1,\textcolor{blue}{0},\textcolor{violet}{0},\textcolor{orange}{1},0,1] \rightarrow [1,\textcolor{orange}{1},\textcolor{violet}{0},\textcolor{blue}{0},0,1]$ & $< L(L-1)/2$\\
    permutation & $[1,\textcolor{blue}{0},\textcolor{violet}{0},\textcolor{orange}{1},\textcolor{teal}{0},\textcolor{magenta}{1}] \rightarrow [1,\textcolor{magenta}{1},\textcolor{orange}{1},\textcolor{blue}{0},\textcolor{violet}{0},\textcolor{teal}{0}]$ & $< \frac{L!}{(\frac{L}{2})!^2}$\\
  \bottomrule
\end{tabular}
\end{center}
\end{table}

\subsection{Selection criterion}

An intuitive first choice for the fitness function would be to use the non-linearity value of the given Boolean function $f$, i.e.:
\begin{equation}
    \text{fitness}_1 = nl_f.
\end{equation}
In \cite{picek2016maximal,picek2017immunological,djurasevic2023digging} a novel and more informative fitness function, which considers an additional property of the Walsh-Hadamard spectrum, is used:
\begin{equation}
    \text{fitness}_2 = nl_f + \frac{2^N-\# maxvalues}{2^N}
\label{fitness2}
\end{equation}
This fitness function has an additional term that penalizes the number of appearances of the maximal absolute value in
the Walsh-Hadamard spectrum, denoted by $\#maxvalues$. The goal is to promote solutions with few occurrences of the maximal absolute value in the spectrum, as these are in a certain phenotypical sense closer to solutions exhibiting higher non-linearity values. This similarity might be demonstrated by inspecting the right tails of the histograms of their absolute spectrum values. The $fitness_2$ function is shown to work particularly well in conjunction with local search (LS) \cite{djurasevic2023digging}. It represents the first step towards considering whole phenotypes when evaluating solutions. Both $fitness_1$ and $fitness_2$ lead to combinatorial (discrete) fitness landscapes, albeit with differing resolutions (degree of graduality).

\begin{figure*}
  \centering
  \includegraphics[width=\linewidth]{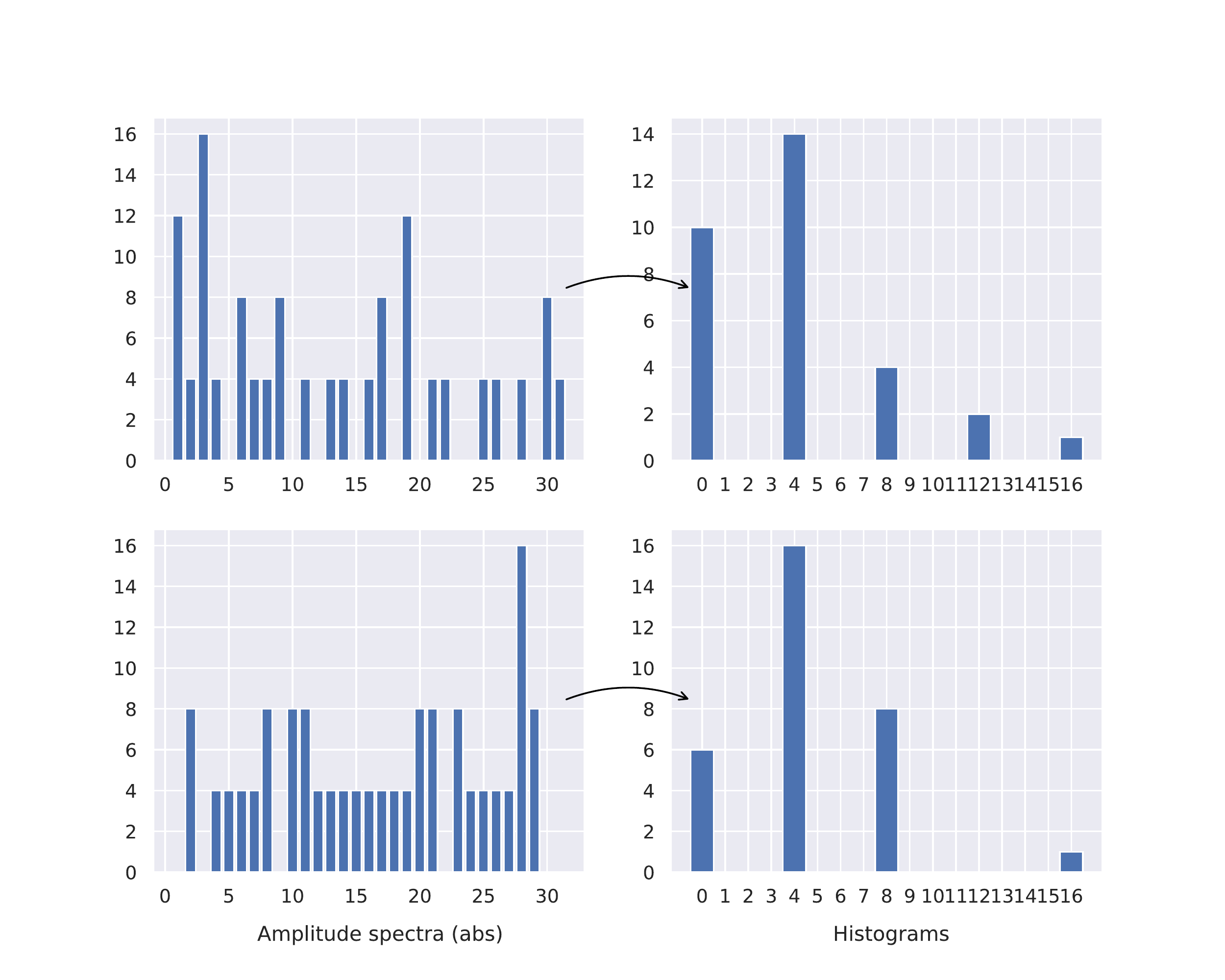}
  \caption{Illustration of the selection criterion. Walsh-Hamadard amplitude spectra (phenotypes) corresponding to two candidate solutions are converted into histograms which are then compared against each other. In the shown example, the non-linearities are equal, as well as the number of appearances of the max value ($16$). The second largest appearing value ($12$) appears once in the spectrum of the first solution and not a single time in that of the second solution. Consequently, the second candidate solution is chosen.}
  \label{Graph1}
\end{figure*}

Building upon such phenotypical considerations, and utilizing even more properties of the spectrum (which represents the phenotype), we propose a novel selection criterion fully inspired by the phenotype,  which can be interpreted as a fully-fledged generalization of $fitness_2$. More specifically, according to our criterion, not only the number of appearances of the maximal absolute value in the spectrum is penalized, but also the number of appearances of the $M$-th largest absolute value, for all $M$. Penalization is less severe for larger $M$ values. Hence, say, a reduction of the number of the maximum absolute value appearances by only one is preferred over a reduction of the number of the second largest absolute value appearances by an arbitrary number. Consequently, when choosing between two solutions, the histograms of their phenotypes (magnitude spectra) are compared. In what follows we formalize the proposed criterion. \\
\textbf{\emph{Selection criterion}}. Given two solutions $x$ and $y$, the one with the higher non-linearity\footnote{Which implies lower maximum absolute value in the spectrum.} is preferred. If the non-linearities are the same, the number of appearances of the largest absolute value in the Walsh-Hadamard spectra are compared, preferring the one with the smaller number of appearances. If it is still a tie, compare the number of appearances of the next ($M$-th) largest possible value in the spectra, $M \in \{2,3, \ldots, M_{max} \}$, in ascending order, and choose the one with the smaller number of appearances. Repeat until the tie is broken. The same procedure is illustrated in the pseudocode provided below in Algorithm \ref{algorithm1} and illustrated graphically in Figure \ref{Graph1}. Observe that, according to the used notation, $\#maxvalues$ is the same as 
$\#larbasval_1$.

\begin{algorithm}
\caption{The proposed selection criterion.}
\label{algorithm1}
\textbf{Inputs:} boolean truth-tables $x$, $y$ \;
\textbf{Output:} selected solution \;
$M\gets 1$\;
calculate \text{WH}($x$) and WH($y$) \tcp*[r]{Calculation of Walsh-Hadamard spectra} 
\eIf{nl($x$) $\neq$ nl($y$)} 
{
    select solution with higher $\text{nl}$\;
}{
    \While{true}
    {
        determine $\#larabsval_M(x)$ and $\#larabsval_M(y)$, numbers of appearances of the $M$-th largest possible value in WH($x$) and WH($y$), respectively\; 
        \If{$\#larabsval_M(x)$  $\neq$ $\#larabsval_M(y)$}
        {select solution with lower $\#larabsval_M$; \\
        break}
        \If{$M=M_{max}$}
        {select solution randomly \\
        break}
        $M\gets M+ 1$\;
    }
}
\end{algorithm}

We raise the point that the criterion boils down to building and comparing the right tails of the histograms of magnitude spectra, starting with the largest value, all until the first difference in the number of appearances is observed. 
Finally, a delicate decision needs to be made as to what happens if the tie is not resolved at all, i.e. if the two resulting histograms are identical. One option is to make the comparison strict, i.e., to only accept strictly better solutions, while the other one is to embrace neutrality. In what follows, for the sake of simplicity, and to avoid the overhead costs associated with accepting a new solution, we choose the former and leave the neutrality considerations for further research. We finally emphasize that instead of using the proposed selection criterion, one might also generalize the fitness function $fitness_2$ by using $M_{max}$ penalization terms instead of merely one. This would result in a fitness function that is discretized on a much finer scale than $fitness_2$ and would in practice require calculations up to impractically large numbers of decimal places when comparing solutions. Hence the use of the proposed selection criterion (as opposed to using it in a form of a fitness function) provides a superior choice in our view. Note that it can also be straightforwardly used in combination with other meta-heuristics, such as GAs and evolutionary strategies (ESs), or, owing to its transitivity, in a slightly modified form to compare more than just two solutions.

\subsection{Local search}

In the context of local search and hill climbers, commonly used in tackling the underlying problem, two main move strategies (pivoting rules) are most frequently used - first-improvement and best-improvement \cite{basseur2015climbing, ochoa2010first, whitley2013greedy}. In the first-improvement variant, as soon as a solution with larger fitness is encountered, it is instantly accepted. On the contrary, in the best-improvement strategy, all neighbors are always evaluated and only then is the one with the largest fitness value selected. In the case of a tie, the next solution can be selected randomly. Although which particular strategy performs better depends on the specifics of the task, it is generally found that a vast majority of landscapes are more effectively explored with the former, while the latter sometimes works better for very smooth landscapes \cite{basseur2015climbing}. Other proposed strategies include, for example, worst and partially worst improvement \cite{basseur2014efficiency}, where the worst improving neighbor among the evaluated neighbors is selected. This again requires evaluation of the whole neighborhood or at least a significant part of it. First-improvement makes for a natural choice in the context of the considered problem given that a swap can increase the non-linearity only by $2$. Hence, searching the remainder of the neighborhood after finding an improving solution would not only be much more costly and inefficient but also potentially useless as no better solution could be found, in the sense of higher non-linearity. Although this may not be necessarily the case if other fitness functions are used, our initial experimentation indicates that best-improvement leads to much larger numbers of required evaluations. On top of that, performing the full neighborhood search for each solution is extremely expensive for larger values of $N$. Consequently, the first-improvement variant is selected.

\section{Experimental results}
\label{chapter:5}

\subsection{Preliminary analysis}

In what follows we perform a preliminary analysis for the sake of selecting the mutation operator. We set $N=8$ ($L=2^N=256$). For each type of mutation, we sample $5000$ genotypes (represented as bitstrings) randomly from the genotypical space, and for each sampled solution, two neighbors are produced by applying the mutation. Then we perform a brief analysis of the fitness correlation between initial solutions and their neighbors, where fitness is simply given by the non-linearity of the solution. The results are summarized in Table \ref{tab:fitnesses}. We employ Spearman's rank correlation coefficient (for measuring the monotonicity of the relationship) and Pearson correlation coefficient (for measuring its linearity). All p-values are significant at $\alpha=0.05$ significance level. As expected, large positive correlation values are seen for all mutations of swap type, with more swaps leading to reduced correlation values. More interestingly, cyclic shifts lead to positive (albeit much smaller) correlation values, despite the fact that this type of mutation leads to large Hamming distances between the original and new genotypes (on average $2^{N-1}$, which is the same as the expected Hamming distance between two randomly and independently generated genotypes). This hints at the fact that a smaller portion of the non-linearity of a Boolean function is not sensitive to such shifts/rotations. Lastly, inversions lead to somewhat larger correlation values than permutations, which might be due to the fact they preserve a part of the structure (like cyclic shifts). 

Taking the results into account, there are several reasons why swaps make for the preferred mutation operator. First, due to high genotypic and phenotypic similarity between neighbors, useful information is not discarded, unlike is the case with other mutation schemes that tend to be more similar to random search. Second, re-computation of the entire Walsh-Hadamard transform is not necessary after each swap, given that there exists a simple update rule that enables evaluation in linear time with respect to the size of the truth table \cite{manzoni2022influence, millan1999boolean}. This significantly reduces complexity and is in line with our primary goal of rendering the search more efficient. Finally, it is characterized by simplicity and yields neighborhoods that are large enough to enable local search to reach solutions with high non-linearity values, especially for smaller $N$.

\begin{table}
  \caption{Neighbours' non-linearity correlations}
  \label{tab:fitnesses}
  \begin{center}
  \begin{tabular}{ccl}
    \toprule
    Mutation&Spearman corr.&Pearson corr.\\
    \midrule
    swap (single) & $0.8638$ & $0.8905$\\
    swap (double) & $0.7628$ & $0.8039$\\
    swap (triple) & $0.6943$  & $0.7438$\\
    cyclic shift & $0.0583$  & $0.0669$\\
    inversion & $0.2779$ & $0.3064$\\
    permutation & $0.2500$ & $0.2645$ \\
  \bottomrule
\end{tabular}
\end{center}
\end{table}

\subsection{Local search results}

Two local search algorithms (one using $fitness_2$ from \cite{djurasevic2023digging}, denoted by \verb|LS-FIT2|, and one relying on the herein introduced selection criterion, called \verb|LS-HISTFIT| referring to histograms) are compared for $6 \leq N \leq 9$ ($2^6 \leq L \leq 2^9$). It should be emphasized that $fitness_2$ provides a strong benchmark, as it was shown to be particularly effective for this problem in previous research \cite{picek2017immunological}, especially in conjunction with local search \cite{djurasevic2023digging}. The results obtained by using $fitness_1$ are not shown given that they are vastly inferior to those resulting from the use of $fitness_2$, which is fully in line with the conclusions drawn in \cite{djurasevic2023digging}. Each run of the local search is performed by starting from a randomly generated perfectly balanced genotype. It proceeds until one of the following happens: a) the target non-linearity value (which depends on the selected size $N$) is obtained, b) convergence to a non-target local optimum takes place, or c) the evaluation budget is exceeded. Convergence (scenario b)) is checked by comparing the current solution with all of its neighbors and is said to happen if no neighbor is better than it. If a) takes place, the run is said to be successful, and otherwise, it is said to have failed. The budget constraint is set to $500000$ evaluations. The target non-linearity values for different $N$ sizes are provided in Table \ref{targetvalues}. For $N \leq 8$, the largest possible or known\footnote{For $n=8$, it is strongly suspected (but not proved) that the maximum non-linearity is $116$, see Dobbertin's conjecture \cite{dobbertin2005construction}. The lowest upper bound is $118$.} values are simply used. However, for $N=9$, the target is set to $236$ because local search most often converged to this value during our initial experimentation. The number of runs is set to $200$ for all $N$ values except in the case $N=9$ when it is set to $25$ due to significant computational expense.

First, consider the percentages of successful runs for the two algorithms and for different $N$ values, given in Table \ref{successfulruns}. For all considered cases, \verb|LS-HISTFIT| yields higher percentages of successful runs. To test for statistical significance, Fisher's exact tests\footnote{Given that the sample sizes are relatively small, while failure probabilities $1-p$ are relatively low, the normality assumption is not fully reasonable, and hence Fisher's exact tests are preferred over one-sided Z tests for proportions.} (for proportions) are performed. Statistical significance at the level of $\alpha = 0.05$ is found for $N=6$ ($p=0.0006$) and $N=8$ ($p=0.0000$), and hence in all these cases, the null hypothesis of the true odds ratio of the populations underlying the observations equals one is rejected. For $N=7$ the obtained p-value is $p=0.12171$ and hence we cannot reject the null hypothesis. Remark that for $N=9$ the target value is smaller than the largest known value ($240$), making successful runs much more likely regardless of the variant used.

\begin{figure*}[t!]
  \centering
  \includegraphics[width=\linewidth]{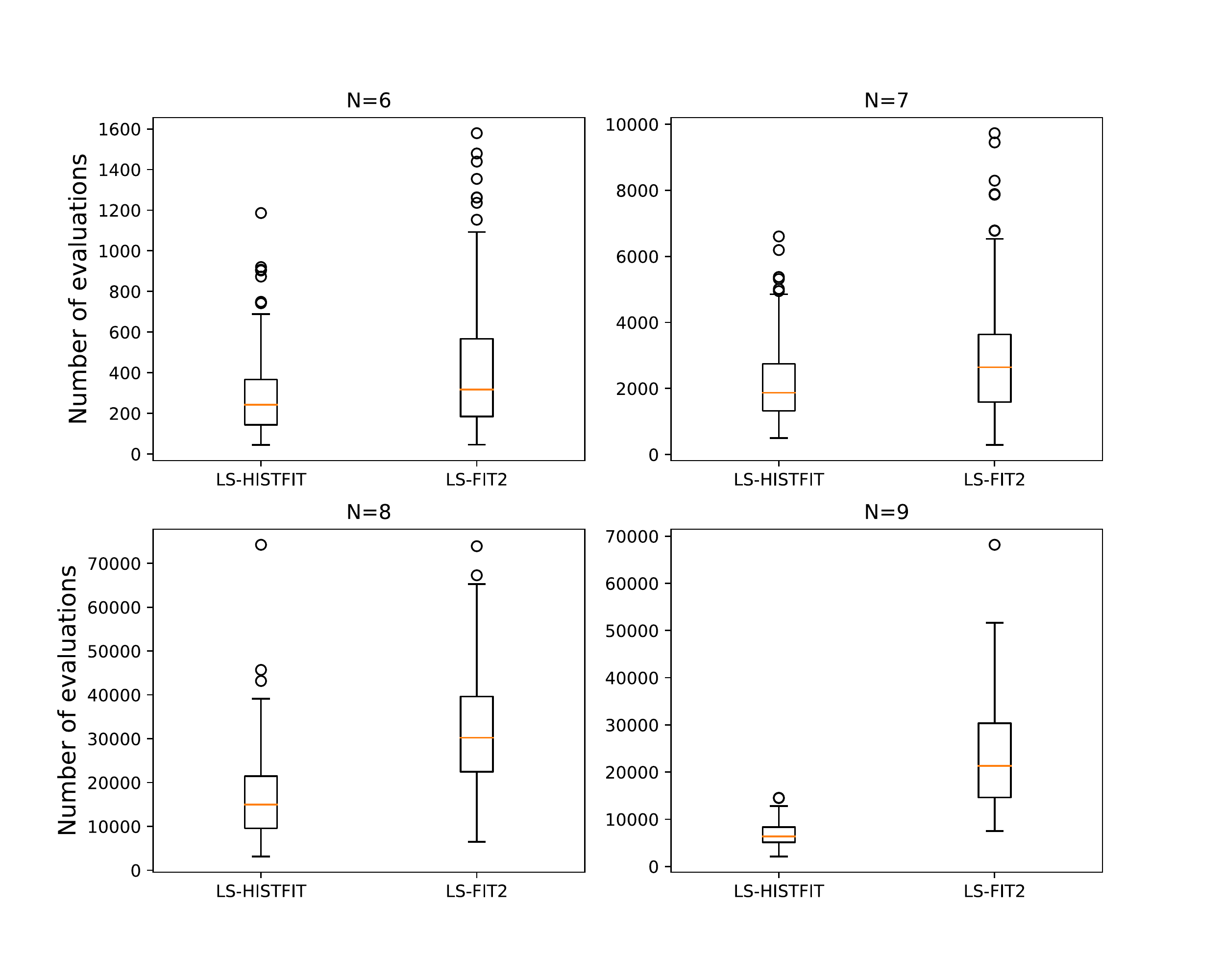}
  \caption{Comparison of convergence speeds (measured via a number of required spectra evaluations) for all consider $N$ values for the two local search variants.}
  \label{Graph2}
\end{figure*}

The main results are shown in Figure \ref{Graph2} and Tables \ref{subtable1} and \ref{subtable2}. Figure \ref{Graph2} depicts the number of fitness evaluations needed to reach the target non-linearity value in successful runs, shown as a box plot together with the respective percentiles. We first note that the results for $N=9$ are better than might be expected because of the previously mentioned reason (laxer target value). In general, the plot clearly indicates that the use of \verb|LS-HISTFIT| selection criterion leads to smaller numbers of required fitness evaluations. To further demonstrate this, we perform the one-sided Mann-Whitney U test on the distributions of the number of evaluating criteria for \verb|LS-HISTFIT| and \verb|LS-FIT2|. The null hypothesis is that there is no significant difference between the distribution of values in the two groups. The alternative hypothesis is that the first distribution is stochastically less than the second distribution. We perform such tests for all considered $N$ values ($6 \leq N \leq 9$). All resulting p-values are less than $10^{-4}$ and hence significant at the $\alpha=0.05$ significance level. Consequently, the null hypothesis is rejected, pointing to the superiority of \verb|LS-HISTFIT| over \verb|LS-FIT2|. Finally, we emphasize that for the \verb|LS-HISTFIT| variant, an additional cost of building the histogram is incurred. However, this involves scattering elements across buckets (histogram bins) which can be done in linear time $\mathcal{O}(L)$.

\begin{table}[htbp]
  \begin{minipage}[t]{0.45\linewidth}
    \centering
    \caption{Target non-linearity values}
    \label{targetvalues}
    \begin{tabular}{cccccl}
      \toprule
      N & Non-linearity value & Remark \\
      \midrule
      6 & $26$ & largest possible value \\
      7 & $56$ & largest possible value \\
      8 & $116$ & largest known value \\
      9 & $236$ & max value obtained in our exp. \\
      \bottomrule
    \end{tabular}
  \end{minipage}%
  \hfill
  \begin{minipage}[t]{0.45\linewidth}
    \centering
    \caption{Percentages of successful runs}
    \label{successfulruns}
    \begin{tabular}{ccc}
      \toprule
      N & Percentage \verb|LS-FIT2| & Percentage \verb|LS-HISTFIT| \\
      \midrule
      $6$ & $91.5 \%$ & $99.0 \%$ \\
      $7$ & $97.0 \%$ & $99.5 \%$ \\
      $8$ & $73.5 \%$ & $97.5 \%$ \\
      $9$ & $100.0 \%$ & $100.0 \%$ \\
      \bottomrule
    \end{tabular}
  \end{minipage}
\end{table}

\begin{table}[htbp]
  \centering
  \begin{minipage}[t]{0.45\linewidth}
    \centering
    \caption{Experimental results (LS-FIT2)}
    \label{subtable1}
    \begin{tabular}{cccccc}
      \toprule
      N & Mean & Std & Median & Min & Max \\
      \midrule
      $6$ & $420.85$ & $324.75$ & $317$ & $46$ & $1579$ \\
      $7$ & $2867.53$ & $1650.74$ & $2642.5$ & $288$ & $9729$ \\
      $8$ & $32168.93$ & $13905.33$ & $30227$ & $6494$ & $73907$ \\
      \midrule
      $9$ & $24081.76$ & $14308.4$ & $21330$ & $7493$ & $68170$ \\
      \bottomrule
    \end{tabular}
  \end{minipage}%
  \hfill
  \begin{minipage}[t]{0.45\linewidth}
    \centering
    \caption{Experimental results (LS-HISTFIT)}
    \label{subtable2}
    \begin{tabular}{cccccc}
      \toprule
      N & Mean & Std & Median & Min & Max \\
      \midrule
      $6$ & $285.29$ & $190.52$ & $242$ & $44$ & $1186$ \\
      $7$ & $2166$ & $1149.85$ & $1871$ & $500$ & $6603$ \\
      $8$ & $16729.99$ & $9299.43$ & $14974$ & $3112$ & $74230$ \\
      \midrule
      $9$ & $7157.6$ & $3215.48$ & $6362$ & $2088$ & $14551$ \\
      \bottomrule
    \end{tabular}
  \end{minipage}
  \label{maintable}
\end{table}

\newpage
\section{Conclusion}
\label{chapter:6}

We proposed a new method for searching for highly non-linear perfectly balanced Boolean functions that utilizes the properties of the phenotypes, represented by Walsh-Hadamard spectra. It is underpinned by a novel selection criterion according to which the phenotypes are directly compared to each other. The experimental results indicate the superiority of the proposed approach with respect to the number of costly Walsh-Hadamard spectrum evaluations that need to be performed to reach the target non-linearity value. In the following work, we plan to investigate the effect of accepting neutral moves in the search for highly non-linear Boolean functions. Some research seems to suggest that in most cases accepting neutral solutions leads to better performance \cite{basseur2015climbing}, although neutrality's relation to the evolutionary computation approaches has been a contentious topic \cite{galvan2011neutrality}. Secondly, we plan to study the feasibility of the methods for larger $N$ values, as well as the use of other encoding schemes besides binary strings (such as trees which might be used in conjunction with genetic programming). Thirdly, a memetic algorithm-based approach might be investigated \cite{ali2009design} by using the local search in combination with a population-based global technique. Finally, another potentially promising path might lie in using novelty search \cite{lehman2011novelty} to try to prevent early convergence by promoting behavioral diversity, or similarly, quality-diversity approaches \cite{pugh2016quality} to obtain a wide repertoire of both phenotypically diverse and high-performing solutions. To this end, alternative phenotypic representations could be employed, such as those speculatively proposed in \cite{picek2015fighting}, like representations that collapse symmetries associated with (balanced) Boolean functions.

\newpage
\bibliographystyle{unsrt}
\bibliography{references}  






\end{document}